\documentclass[runningheads]{llncs}
\usepackage{graphicx}
\usepackage{rotating}

\usepackage{tikz}
\usepackage{comment}
\usepackage{amsmath,amssymb} 
\usepackage{color}
\usepackage{bm}

\usepackage{booktabs}
\usepackage{multirow}
\usepackage{wrapfig}
\usepackage{subfigure}

\usepackage[accsupp]{axessibility}  

\definecolor{citecolor}{RGB}{65,105,225}
\usepackage[pagebackref=true,breaklinks=true,letterpaper=true,colorlinks,
citecolor=citecolor,bookmarks=true]{hyperref}

\usepackage{xcolor}
\usepackage{marvosym}

\begin{document}
\pagestyle{headings}
\mainmatter
\def\ECCVSubNumber{2378}  

\title{StyleLight: HDR Panorama Generation for Lighting Estimation and Editing} 

\titlerunning{StyleLight}
%



\author{Guangcong Wang \and
Yinuo Yang \and
Chen Change Loy \and
Ziwei Liu\textsuperscript{\Letter}} 

\authorrunning{G. Wang et al.}
\institute{S-Lab, Nanyang Technological University \\
\email{\{guangcong.wang,ccloy,ziwei.liu\}@ntu.edu.sg \quad YANG0689@e.ntu.edu.sg}}

\maketitle

\begin{figure}
    \centering
  \includegraphics[width=\linewidth]{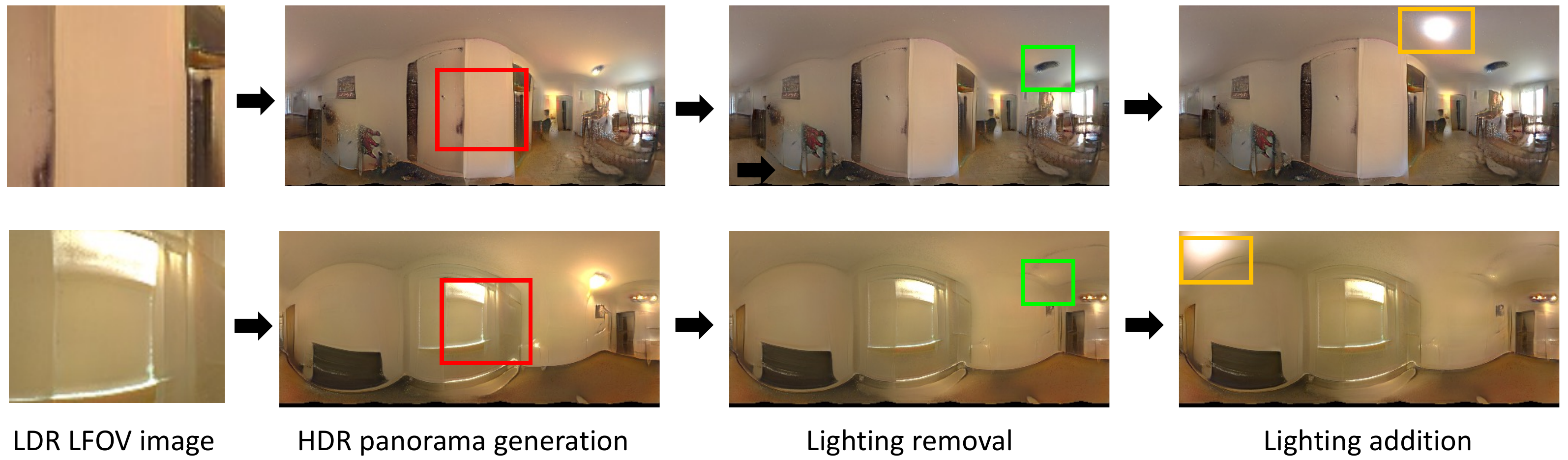}
  \caption{\textbf{StyleLight.} The proposed StyleLight estimates HDR panorama lighting from an LDR LFOV image and performs lighting addition and removal.}
  \label{fig:teaser}
\end{figure}

\begin{abstract}
We present a new lighting estimation and editing framework to generate high-dynamic-range (HDR) indoor panorama lighting from a single limited field-of-view (LFOV) image captured by low-dynamic-range (LDR) cameras. Existing lighting estimation methods either directly regress lighting representation parameters or decompose this problem into LFOV-to-panorama and LDR-to-HDR lighting generation sub-tasks. However, due to the partial observation, the high-dynamic-range lighting, and the intrinsic ambiguity of a scene, lighting estimation remains a challenging task. To tackle this problem, we propose a coupled dual-StyleGAN panorama synthesis network (\textbf{StyleLight}) that integrates LDR and HDR panorama synthesis into a unified framework. The LDR and HDR panorama synthesis share a similar generator but have separate discriminators. During inference, given an LDR LFOV image, we propose a focal-masked GAN inversion method to find its latent code by the LDR panorama synthesis branch and then synthesize the HDR panorama by the HDR panorama synthesis branch. StyleLight takes LFOV-to-panorama and LDR-to-HDR lighting generation into a unified framework and thus greatly improves lighting estimation. Extensive experiments demonstrate that our framework achieves superior performance over state-of-the-art methods on indoor lighting estimation. Notably, StyleLight also enables intuitive lighting editing on indoor HDR panoramas, which is suitable for real-world applications. Code is available at  \url{https://style-light.github.io/}.


\end{abstract}

\section{Introduction}
Environmental lighting models aim to approximate realistic lighting effects in a scene. Existing state-of-the-art methods \cite{gardner2017learning,kanamori2019relighting,pandey2021total,sun2019single} use HDR spherical panoramas for lighting. In many real-world applications, however, only LDR LFOV images are available. Therefore, it is important to estimate HDR panorama illumination from an LDR LFOV image, aided with the capability of lighting editing to flexibly control lighting conditions. Lighting estimation and editing have a wide range of applications in mixed reality such as object insertion and relighting in virtual meetings and games. For example, given an LDR LFOV background, a person can be inserted into the background and realistically relit by controllable panorama illumination.

There are three main challenging problems in lighting estimation. 
First, only a partial scene is observed in a limited FOV image. The other unobserved part of a scene is ambiguous and could significantly affect the illumination distribution in a full scene due to ambiguous lighting sources with different numbers, shapes, and directions.
Second, only a partial range of lighting is observed in typical LDR images. Physical lighting has a wide range of intensity levels in real scenes, ranging from shadows to bright light sources. However, in LDR images, underexposed areas are clipped to 0, and overexposed areas are clipped to 1. Out-of-range lighting cannot be observed in LDR images. 
Third, lighting estimation of unobserved views depends on intrinsic attributes of observed objects (e.g., material and 3D geometry), but these annotations are often unavailable in real-world applications.

Existing methods on illumination estimation provide partial solutions to illumination estimation, which could be roughly classified into two groups, i.e., regression-based estimation methods \cite{gardner2017learning,gardner2019deep,garon2019fast,zhan2021sparse,hold2017deep,legendre2019deeplight} and generation-based estimation methods \cite{zhan2020emlight,song2019neural,srinivasan2020lighthouse}. 
In the first group, researchers propose to directly regress panorama lighting information from a given LDR LFOV image. The main difficulty of this pipeline is to parameterize lighting representations as target labels for regression, including light representation (e.g., number, shape, directions and intensity) \cite{gardner2019deep}, Spherical Harmonics \cite{garon2019fast}, and wavelet transformation \cite{zhan2021sparse}. However, these lighting labels mainly focus on strong lighting sources and thus lose the details of a scene. \textcolor{black}{When a scene contains reflective materials (e.g., glass, mirror, and metal), this pipeline will fail to provide a desired rendered visual effect. In image-based rendering, the HDR map can be also applied as background.} Moreover, regression-based models often suffer from unstable training of neural networks due to partial observed view, a wide range of lighting, unknown geometry, and material of objects. For example, EMlight \cite{zhan2020emlight} claimed that it is necessary to train a regression model by gradually increasing training examples to avoid training collapse.  
Second, generation-based estimation methods aim at generating full HDR panoramas from an LDR LFOV image. They typically decompose illumination estimation as several sub-tasks, such as LFOV-to-panorama completion, LDR-to-HDR regression \cite{zhan2020emlight}, and 3D geometry prediction \cite{song2019neural}. \textcolor{black}{These methods either require additional annotations or multiple optimization stages.}

Our proposed method belongs to the second group. Unlike the existing \sloppy{generation-based} estimation methods, we propose a coupled dual-StyleGAN panorama synthesis network (\textbf{StyleLight}) that solves LFOV-to-panorama and LDR-to-HDR problems in a unified framework and does not require any 3D geometry annotation. Specifically, StyleLight is a coupled dual-StylGAN that maps noises to both HDR and LDR panoramas. The LDR and HDR panorama synthesis share a similar generator. At the output of the generator, a tone mapping is adopted for the translation between HDR and LDR. Considering the fact that an HDR panorama consists of extremely different weak lighting distribution and strong lighting distribution, we use two discriminators to distinguish real/fake HDR and LDR panorama distributions, respectively. Given an LDR LFOV image, we propose a focal-masked GAN inversion method to find its latent code through the LDR panorama synthesis branch and then generate its corresponding HDR panorama as illumination maps through the HDR synthesis branch. Moreover, we propose a structure-preserved GAN inversion method for lighting editing with a trained StyleLight model to flexibly control the environment lighting of panoramas, such as turning on/off lights, opening/closing doors and windows, making our method well-suited for various real-world applications.

The main contributions of this paper can be summarized as follows:
\textbf{1)} We propose a coupled dual-styleGAN synthesis network (StyleLight) that integrates HDR and LDR panorama synthesis in a unified framework. During inference, we propose a focal-masked GAN inversion method to solve both LFOV-to-panorama and LDR-to-HDR generation. 
\textbf{2)} We propose a structure-preserved GAN inversion method for lighting editing with the trained StyleLight model to flexibly control panorama lighting. 
\textbf{3)} Extensive experiments demonstrate the superiority and the effectiveness of our proposed lighting estimation method over state-of-the-art methods on indoor HDR panoramas and show promising applications with our proposed lighting editing.

\section{Related Work}
Lighting estimation is one of the long-standing problems in computer vision and graphics. Early work \cite{barron2013intrinsic,wu2011high,valgaerts2012lightweight} typically simplifies the models by considering fixed number of lights \cite{lopez2010compositing}, annotated light positions \cite{karsch2011rendering} and geometry \cite{karsch2011rendering,karsch2011rendering,lombardi2015reflectance,karsch2011rendering,maier2017intrinsic3d,maier2017intrinsic3d}. Recent lighting estimation methods relax such assumptions and focus on more challenging lighting estimation. 
We briefly review the work most relevant to this paper.

\noindent
\textbf{Lighting estimation.} Existing lighting estimation methods can be roughly classified into two groups, i.e., regression-based methods and generation-based methods. Regression-based models \cite{gardner2019deep,legendre2019deeplight,garon2019fast,zhan2021sparse,zhan2020emlight,hold2017deep} seek for effective lighting representations of HDR panoramas as labels for lighting regression. These methods take an LDR LFOV image as input and predict lighting labels, such as direct lighting representation (e.g., numbers, shapes, directions, and intensity) \cite{gardner2019deep}, Spherical Harmonics (SH) \cite{garon2019fast}, and wavelet transformation \cite{zhan2021sparse}. For example, 
Gardner \textit{et al.} \cite{gardner2017learning} trained a lighting classifier on an annotated LDR panorama dataset to predict the location of lighting sources of LDR LFOV images and fine-tuned the prediction of lighting intensity on a small HDR panorama dataset.
Gardner \textit{et al.} \cite{gardner2019deep} regressed the illumination conditions such as the light source positions in 3D, areas, intensities, and colors.
LeGendre \textit{et al.} \cite{legendre2019deeplight} designed a joint camera-sphere device to collect LDR LFOV images that include three reflective spheres, and regressed HDR lighting from an LDR LFOV image.
Garon \textit{et al.} \cite{garon2019fast} regressed the 5th-order SH coefficients for the lighting at a certain location given a single image and a 2D location in an image. Regression-based models ignore the low-intensity lighting details and thus fail to provide a desired rendered quality for mirror materials. Besides, they often suffer from unstable training of neural networks due to the high dynamic range of lighting.

Generation-based methods \cite{gardner2017learning,song2019neural,srinivasan2020lighthouse} focus on decomposing illumination estimation as several sub-tasks, such as LFOV-to-panorama completion and LDR-to-HDR regression. 
Song \textit{et al.} \cite{song2019neural} decomposed illuminations prediction into several simpler differentiable sub-tasks including geometry estimation, scene completion, and LDR-to-HDR estimation. Zhan \textit{et al.} \cite{zhan2020emlight} decomposed the illumination map into spherical light distribution, light intensity and the ambient term for illumination regression. This aforementioned paradigm either decomposes illumination estimation as several sub-tasks or collects expensive annotations for training.

\noindent
\textbf{Image editing via GAN inversion.} Image editing is a task that modifies a target attribution of a given image while preserving other details \cite{wang2021high,harkonen2020ganspace,li2021transforming,shen2020interpreting,roich2021pivotal,zhuang2021enjoy,tewari2020stylerig,shen2021closed}. Recent state-of-the-art image editing methods focus on generative adversarial network (GAN) inversion techniques to control the latent space of GANs for image editing. 
For example, Ganspace \cite{harkonen2020ganspace} used Principal Component Analysis (PCA) to identify important latent directions and create interpretable controls for image attributes, such as viewpoint, aging, lighting, and time of day. 
InterfaceGAN \cite{shen2020interpreting} proposed to edit the latent code $\mathbf{z}$ with $\mathbf{z}_{edit} = \mathbf{z}+\alpha \mathbf{n}$ for each semantic attribute where $\mathbf{n}$ is a normal vector of a linear SVM's hyper-plane. 
SeFa \cite{shen2021closed} proposed a closed-form factorization algorithm for latent semantic discovery by directly decomposing the pre-trained weights. 
StyleFlow \cite{abdal2021styleflow} formulated conditional exploration as an instance of conditional continuous normalizing flows in the GAN latent space conditioned by attribute features. 
These image editing methods focus on controlling the latent code to control attributes of objects. However, in our lighting editing setting, our goal is to edit a light source that is located at a small area of a panorama given a bounding box, which is different from the typical image editing.

\section{Our Approach}

\subsection{Overview of the Proposed Framework}
\label{subsec:overview}
\begin{figure*}[t]
  \centering
  \includegraphics[width=\linewidth]{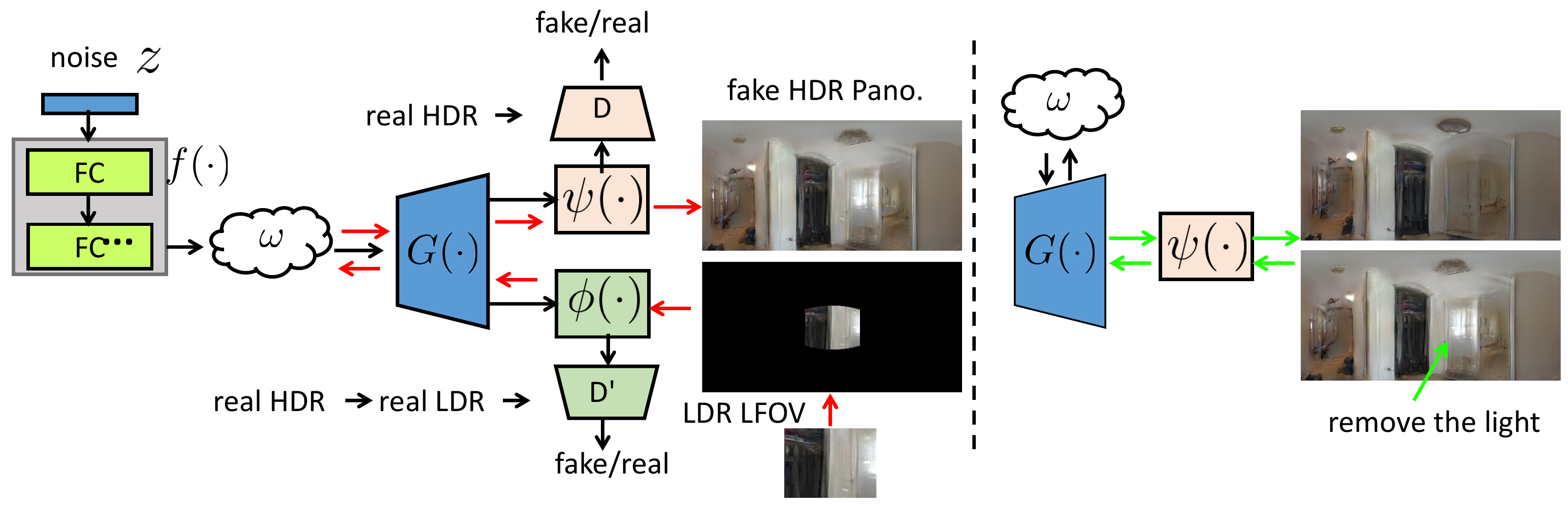}
  \caption{Overview of the proposed framework. \textbf{Left:} Lighting estimation. We first propose a coupled dual-StyleGAN that generates HDR and LDR panoramas from noises. Given an LDR LFOV image, we map it into a panorama. We introduce a focal-masked GAN inversion method to find its latent code through the LDR branch and compute the HDR panorama through the HDR branch. \textbf{Right:} Lighting editing. With the trained dual-StyleGAN, we perform lighting editing to control panoramas lighting.}\label{fig:overview}
\end{figure*}
An overview of our proposed framework is depicted in \textbf{Fig. \ref{fig:overview}}. The framework consists of lighting estimation and lighting editing. The goal of lighting estimation is to predict HDR panorama from an LDR LFOV image. To achieve this goal, we introduce a novel coupled dual-StyleGAN synthesis network (StyleLight) that learns to synthesize both HDR and LDR panoramas from noises with a shared generator and two specific discriminators. With the trained StyleLight model, we propose a focal-masked GAN inversion method to find the latent code of an LDR LFOV image and predict the HDR panorama as an illumination map (Section \ref{subsec:lighting_estimation}). To control various lighting conditions of the target HDR panorama, we introduce a structure-preserved lighting editing method to solve this problem (Section \ref{subsec:light_edit}). To the best of our knowledge, this is the first attempt to perform lighting editing on HDR panoramas for controlling the scene lighting.

\subsection{Lighting Estimation}
\label{subsec:lighting_estimation}
Generation-based lighting estimation methods typically decompose illumination estimation into several sub-tasks, such as LFOV-to-panorama completion and LDR-to-HDR regression. These methods optimize sub-tasks independently and could lead to sub-optimal estimation results. Moreover, these methods can be hardly scalable to lighting editing. Different from these methods, we propose a coupled dual-StyleGAN that integrates LFOV-to-panorama, LDR-to-HDR, and lighting editing into a unified framework.

\begin{figure*}[ht]
  \centering
  \includegraphics[width=\linewidth]{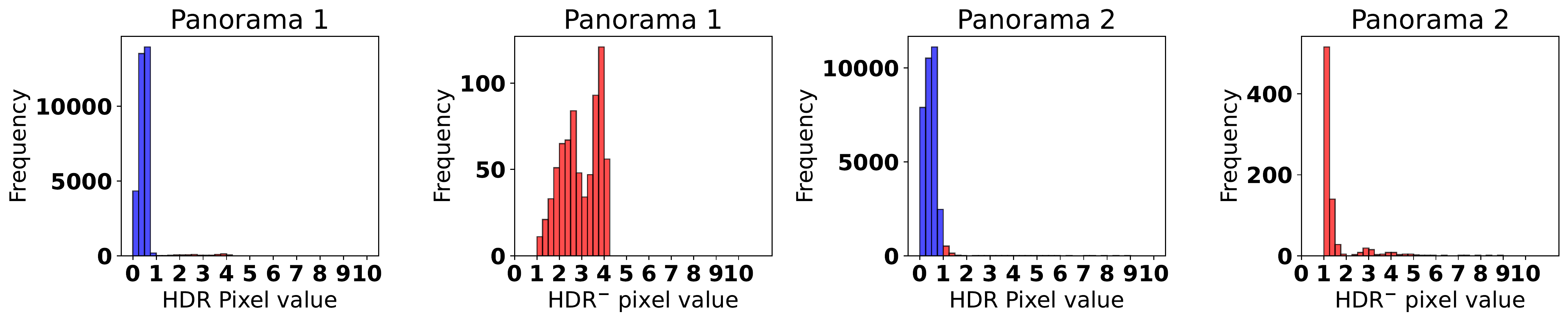}
  \caption{Pixel distributions of two HDR panoramas. The first and third histograms cover the full range of pixel values. The second and fourth histograms visualize high-intensity values of the first and third histograms. An HDR panorama consists of two components, i.e., low-intensity pixels (blue, from 0 to 1, denoted as LDR) and high-intensity pixels that are out of range LDR pixels (red, larger than 1, denoted as HDR$^{-}$). }\label{fig:pixel_distribution}
\end{figure*}

\noindent
\textbf{Coupled dual-StyleGAN for joint LDR and HDR panorama synthesis.} HDR panoramas are more difficult to synthesize than LDR panoramas due to the complex characteristics of distributions. \textbf{Figure \ref{fig:pixel_distribution}} shows the distributions of two HDR panorama examples (others are similar). An HDR panorama consists of two components, i.e., low-intensity pixels (blue, denoted as LDR, from 0 to 1) and high-intensity pixels that are out of range LDR pixels (red, denoted as HDR$^{-}$, larger than 1). The distributions suggest three key points. First, LDR distributions are similar to a Gaussian distribution. Most of the pixels are LDR pixels and they are limited in the narrow range. This suggests that LDR distributions are easy to generate. Second, HDR$^{-}$ distributions are dissimilar and cover a few pixels in the high dynamic range (much larger than 1). This implies that HDR$^{-}$ distributions are difficult to generate and could lead to unstable training of HDR synthesis. If we only use an HDR discriminator to distinguish entire HDR panoramas, \textcolor{black}{the dynamic HDR$^{-}$ pixels (e.g., $>$100) largely affect the discriminator and other regions (0$\sim$1) are often ignored through convolution. $D_{'}$ can be regarded as a regularization term to preserve image semantics in LDR regions.} \textcolor{black}{We analyze the distributions of 1401 training images. We fit a Gaussian distribution for both LDR and HDR$^-$ of each image, respectively. We find that most of LDR distributions (99.86\% LDR images) follow Gaussian distributions and most of HDR$^-$ distributions (70.38\% HDR$^-$ images) fail to converge to Gaussian distributions.} Note that LDR panoramas play a key role in the synthesis from LDR LFOV images to HDR panoramas in the next section. We expect HDR and LDR images to be well synthesized. 
It is noteworthy that although HDR$^{-}$ distributions are difficult to generate, HDR$^{-}$ and LDR have a strong semantic relation. LDR pixels provide the location and shape contextual information of lighting sources while HDR pixels provide intensity information.

Motivated by the fact that LDR panoramas are easy to generate and they provide location and shape contextual information of lighting for HDR synthesis, we introduce a coupled dual-StyleGAN for joint HDR and LDR panorama synthesis. With the help of LDR panorama synthesis, dual-StyleGAN is able to generate accurate HDR panoramas. Specifically, the dual-StyleGAN takes a noise vector $\bm{z}\in z$ as input and maps the noise ${\bm{z}}$ into a intermediate latent space $\omega$ that is much less entangled than the $z$ space, as analyzed in StyleGAN2 \cite{karras2020analyzing}. The mapping function $\bm{w}=f({\bm{z}})$ consists of eight fully-connected layers with activation layers. The latent code $\bm{w}\in \omega$ is then fed into the generator $G$. After that, two transformations $\psi(\cdot)$ and $\phi(\cdot)$ are used to transform the output of $G$ into HDR and LDR panoramas, respectively. That is, $x_{\mathrm{hdr}}=\psi(G({\bm{w}}))$ and $x_{\mathrm{ldr}}=\phi(G({\bm{w}}))$, as illustrated in \textbf{Fig. \ref{fig:overview}}. \textcolor{black}{The HDR output $\psi(\cdot)$ is an inverse tone-mapping plus clamping (positive pixels) and the LDR output $\phi(\cdot)$ is clampped from 0 to 1. To ease the stable training, we pre-process the the HDR training panoramas by $x^{'}=\alpha x^{1/{\gamma}}$, where $\gamma=2.4$ and $\alpha$ is set to 0.5 divided by 50-th percentile of all pixel values. We use $clamp(x^{'},0,+\infty))$ for HDR and $clamp(x^{'},0,1)$ for LDR. During inference, we perform inverse tone-mapping $clamp(x^{'},0,+\infty))$ into $x$. Other tone-mappings are also suitable for our framework.} The idea behind the shared generator network is that HDR and LDR panoramas share the same contextual semantics. Formally, the proposed coupled dual-StyleGAN is formulated as
\begin{equation}
\label{eq:hdr_ldr}
\begin{aligned}
 \mathop {\min }\limits_{G} \mathop {\max }\limits_{D,D^{'}}\mathcal{J}(G,D,D^{'}) &= E_{x \backsim P_{data}(x)}[\log(D(x))] \\
   &+E_{x \backsim P^{'}_{data}(x)}[\log(D^{'}(x))]\\
   &+E_{w\backsim P_{\bm{w}}}[\log(1-D(\psi(G(\bm{w}))))]\\
   &+E_{w\backsim P_{\bm{w}}}[\log(1-D^{'}(\phi(G(\bm{w}))))],\\
\end{aligned}
\end{equation}
where $D(\cdot)$ and $D^{'}(\cdot)$ denote two discriminators for HDR synthesis and LDR synthesis, respectively. \textcolor{black}{In Eq. \ref{eq:hdr_ldr}, we optimize $G$ by minimizing the third and fourth terms. We optimize $D$ and $D^{'}$ simultaneously by maximizing the probability of predicting real and fake panoramas.} Note that we do not use the same discriminator because the distributions of HDR and LDR panoramas are different, as illustrated in \textbf{Fig. \ref{fig:pixel_distribution}}. 

\noindent
\textbf{Focal-Masked GAN inversion for LDR-to-HDR transformation and LFOV-to-panorama completion.} After obtaining the coupled dual-StyleGAN, we can perform LDR-to-HDR transformation and LFOV-to-panorama completion. In the coupled dual-StyleGAN, HDR synthesis and LDR synthesis share the same generator $G$ and thus guarantee pixel-level alignment between the LDR-to-HDR transformation. Thanks to the coupled dual-StyleGAN that integrates HDR synthesis and LDR synthesis into a unified framework, we can find a shared latent code of an HDR panorama and an LDR LFOV image in the $\omega$ space, as illustrated in \textbf{Fig. \ref{fig:overview}}(left). Given an LDR LFOV image, we map it into a spherical panorama as a masked panorama. We then extend a GAN inversion method \cite{roich2021pivotal} into a masked GAN inversion to tailor for our coupled dual-StyleGAN. We find the latent code of the masked panorama in the $\omega$ space. We forward the latent code through the HDR branch to obtain an HDR panorama. Let $x_{\mathrm{lfov}}^{\mathrm{ldr}}$ denote a masked panorama of the LDR LFOV image and $\bm{n}$ denote noise maps of all layers. We first optimize latent code $\bm{w}\in \omega$ such that synthesized LDR LFOV matches the target LDR LFOV as much as possible, which can be formulated as 
\begin{equation}
\label{eq:inversion}
  \bm{w}_{*}, \bm{n}_{*}=\mathop{\mathrm{argmin}}\limits_{\bm{w},\bm{n}}  \mathcal{L}_{\mathrm{LPIPS}}(M\odot\phi(G(\bm{w},\bm{n};\theta)),x_{\mathrm{lfov}}^{\mathrm{ldr}})+ \lambda_{\bm{n}}\mathcal{L}_{\bm{n}}(\bm{n}),
\end{equation}
where $\mathcal{L}_{\mathrm{LPIPS}}$ represents the VGGNet based Perceptual loss \cite{johnson2016perceptual} for distance measure and $\mathcal{L}_{\bm{n}}(\bm{n})$ represents a noise regularization term to constrain noise maps $\bm{n}$ and $\lambda_{\bm{n}}$ controls its importance. We re-use the symbol $G$ with latent code $\bm{w}\in \omega$, noise maps $\bm{n}$ and weights $\theta$ as input. The variable $M$ denotes a mask that preserves the pixels corresponding to the LFOV image $x_{\mathrm{lfov}}^{\mathrm{ldr}}$. Because it is hard to directly reconstruct the target LDR LFOV image by optimizing $\bm{w}$ and $\bm{n}$, we then fine-tune the weights $\theta$ of G by 
\begin{equation}
 \begin{aligned}
 \label{eq:gan_finetue}
 \mathcal{L}_{G}(\theta)=&  \mathcal{L}_{\mathrm{LPIPS}}(M\odot\phi(G(\bm{w}_{*},\bm{n}_{*};\theta)),x_{\mathrm{lfov}}^{\mathrm{ldr}})\\
 &+\lambda_{L2}^{R} \mathcal{L}_{L2}(M\odot\phi(G(\bm{w}_{*},\bm{n}_{*};\theta)),x_{\mathrm{lfov}}^{\mathrm{ldr}}),\\
 \end{aligned}
\end{equation}
 where $\mathcal{L}_{G}$ represents the similarity loss between a masked generated LDR image and the LFOV HDR image $x_{\mathrm{lfov}}^{\mathrm{ldr}}$. We use the Mean-Squared-Error loss and the VGGNet based Perceptual loss for distance measure. To further improve visual quality, we also use a regularization term, which is given by 
 \begin{equation}
 \begin{aligned}
\label{eq:gan_regularization}
 \mathcal{L}_{R}(\theta)&=  \mathcal{L}_{\mathrm{LPIPS}}(\phi(G(\bm{w}_r,\bm{n}_{*};\theta)),\phi(G(\bm{w}_r,\bm{n}_{*};\theta_0)))\\&+ \lambda_{L2}^{R'}\mathcal{L}_{L2}^{R}(\phi(G(\bm{w}_r,\bm{n}_{*};\theta)),\phi(G(\bm{w}_r,\bm{n}_{*};\theta_0))),\\
 \end{aligned}
\end{equation}
where $\bm{w}_r=\bm{w}_{*}+\alpha(\bm{w}_z-\bm{w}_{*})/||\bm{w}_z-\bm{w}_{*}||_{2}$ and $\bm{w}_z$ is a latent code that is generated by a random $\bm{z}$. The $\theta_0$ is the old weights of G before finetuning. The $\alpha$ is a interpolation parameter. Finally, we combine Equations \ref{eq:gan_finetue} and \ref{eq:gan_regularization}, and fine-tune $G$ by 
 \begin{equation}
\label{eq:final_loss}
 \theta_{*}=  \mathop{\mathrm{argmin}}\limits_{\theta}\mathcal{L}_{G}(\theta)+\eta\mathcal{L}_{R}(\theta).
\end{equation}

Simply extending the GAN inversion method \cite{roich2021pivotal} into a masked GAN inversion for the coupled dual-StyleGAN cannot guarantee that the pixels of lights in LDR panoramas will be transformed into high-intensity HDR$^{-}$. Strong lighting sources are the main factors for rendering environment, but high-intensity pixels are much less than low-intensity pixels, as shown in \textbf{{Fig. \ref{fig:pixel_distribution}}}. Without any constraint, the model tends to fit the major LDR pixels and ignore the few HDR$^{-}$ pixels. Therefore, we propose a focal-masked to highlight the strong lighting sources. In our setting, we highlight top 10\% strongest pixels. Given an LDR LFOV image $x_{\mathrm{lfov}}^{\mathrm{ldr}}$, we compute the top 10\% strongest pixels and obtain a focal mask, which denoted as $FOCAL(x_{\mathrm{lfov}}^{\mathrm{ldr}})$. We re-write Equation \ref{eq:inversion} as 
\begin{equation}
\label{eq:inversion_new}
\begin{aligned}
  \bm{w}_{*}, \bm{n}_{*}&=\mathop{\mathrm{argmin}}\limits_{\bm{w},\bm{n}}  \mathcal{L}_{\mathrm{LPIPS}}(M\odot\phi(G(\bm{w},\bm{n};\theta)),x_{\mathrm{lfov}}^{\mathrm{ldr}})+ \lambda_{\bm{n}}\mathcal{L}_{\bm{n}}(\bm{n})\\
  &+\beta_{L2}\mathcal{L}_{L2}(FOCAL(\phi(G(\bm{w},\bm{n};\theta))),FOCAL(x_{\mathrm{lfov}}^{\mathrm{ldr}})).\\
\end{aligned}
\end{equation}
Note that $FOCAL(\phi(G(\bm{w},\bm{n};\theta)))$ is consistent with $FOCAL(x_{\mathrm{lfov}}^{\mathrm{ldr}})$ in Equation \ref{eq:inversion_new}. Similarly, we re-write Equation \ref{eq:gan_finetue}  as
\begin{equation}
\label{eq:gan_finetune_new}
\begin{aligned}
   \mathcal{L}_{G}(\theta)&=  \mathcal{L}_{\mathrm{LPIPS}}(M\odot\phi(G(\bm{w}_{*},\bm{n}_{*};\theta)),x_{\mathrm{lfov}}^{\mathrm{ldr}})\\
   &+\lambda_{L2}^{R} \mathcal{L}_{L2}(M\odot\phi(G(\bm{w}_{*},\bm{n}_{*};\theta)),x_{\mathrm{lfov}}^{\mathrm{ldr}})\\
   &+\beta_{L2}\mathcal{L}_{L2}(FOCAL(\phi(G(\bm{w}_{*},\bm{n};\theta))),FOCAL(\bm{w}_{\mathrm{lfov}}^{\mathrm{ldr}})).\\
\end{aligned}
\end{equation}

\subsection{Lighting Editing}
\label{subsec:light_edit}
Lighting estimation provides a way to replicate real-world lighting conditions of a scene from a single LDR LFOV image and can be used to re-render inserted objects. In some occasions, however, the estimated lighting conditions do not meet our design/modeling requirements. For example, we would like to turn on an indoor light and ``turn off" the light from windows at night given daytime HDR panoramas. Although light sources are located in a local area, it affects the global panorama lighting. Existing approaches do not consider this problem. Here, we introduce a lighting editing method to control panorama lighting through exploiting the powerful StyleLight.

In general, lighting editing includes three operations, i.e, adding a new light, removing a light, and controlling lighting intensity. In essence, the first two operations can be accomplished by the third one.  In light estimation, we obtain a latent code $w_0$ that recovers an HDR panorama from an LDR LFOV image. Here we assume that $\bm{w}_0$ is known. Our goal is to find a editable direction of $\bm{w}_0$ to adjust the lighting intensity such that the lighting intensity at a candidate  location changes while the geometry structure of the scene is preserved. Given a bounding box of a candidate lighting location as a mask $M$, we introduce a structure-preserved lighting editing loss to find the editable direction of $\bm{w}$, which is formulated as
\begin{equation}
\label{eq:inversion_light}
\begin{aligned}
  \bm{w}_{*}, \bm{n}_{*}&=\mathop{\mathrm{argmin}}\limits_{\bm{w},\bm{n}}  \mathcal{L}_{\mathrm{LPIPS}}((1-M)\odot \phi(G(\bm{w},\bm{n};\theta)),(1-M)\odot\phi(G(\bm{w}_0,\bm{n};\theta))\\
  &+\mathcal{L}_{L2}((1-M)\odot \phi(G(\bm{w},\bm{n})),(1-M)\odot\phi(G(\bm{w}_0,\bm{n};\theta))\\
  &+ \frac{\delta}{N}\sum M\odot\phi(G(\bm{w},\bm{n};\theta)),\\
\end{aligned}
\end{equation}
where $\delta$ denotes a lighting adjusting factor. When $\delta$ is positive, the lighting intensity decreases; when $\delta$ is negative, the lighting intensity increases. $N$ is the number of masked pixels. \textcolor{black}{The third term aims to adjust lighting intensity given a bounding box, and the first and second terms preserve other regions.}

\section{Experiments}
\label{sec:exper}
\textbf{Datasets and experimental settings.} We evaluate our StyleLight model on the Laval Indoor HDR dataset \cite{gardner2017learning} that contains 2,100 HDR panoramas. Following the train/test split of \cite{gardner2017learning}, we use 1,719 panoramas for training and 289 panoramas for testing. For the test set, we crop one HDR LFOV image from each panorama and transform it into an LDR LFOV image by tone mapping \cite{zhan2020emlight}. Since lighting estimation is a challenging task, some compared methods use extra annotations. For example, Gardner2017 \cite{gardner2017learning} used annotated lighting sources of SUN360 \cite{xiao2012recognizing} to learn to predict the locations and shapes of lighting sources. Gardner2019 \cite{gardner2019deep} manually annotated the Laval Indoor dataset using EnvyDepth \cite{banterle2013envydepth} and used per-pixel depth maps to improve lighting estimation.

\noindent
\textbf{Implementation details.} 
We use StyleGAN2 \cite{karras2020analyzing} as the backbone. The output of the generator is a 256$\times$512 image. We resize it into 128$\times$256 for evaluation, which is consistent with other methods. The resolution of feature maps begins with a $4\times 8$ constant map and increases by $\times 2$ upsampling each block, yielding $256\times 512$ patch-based feature maps. In the full training of StyleLight, 4000k real images (kimg) are shown to the discriminators. We train StyleLight with 8 v100 GPUs for 7 hours. The batch size is set to 32. We use Adam optimizer for the generators and two discriminators with a learning rate of 0.0025 and set $\beta_1=0$,  $\beta_2=0.99$.
\begin{wrapfigure}{r}{0.4\textwidth}
  \centering
    \includegraphics[width=0.4\textwidth]{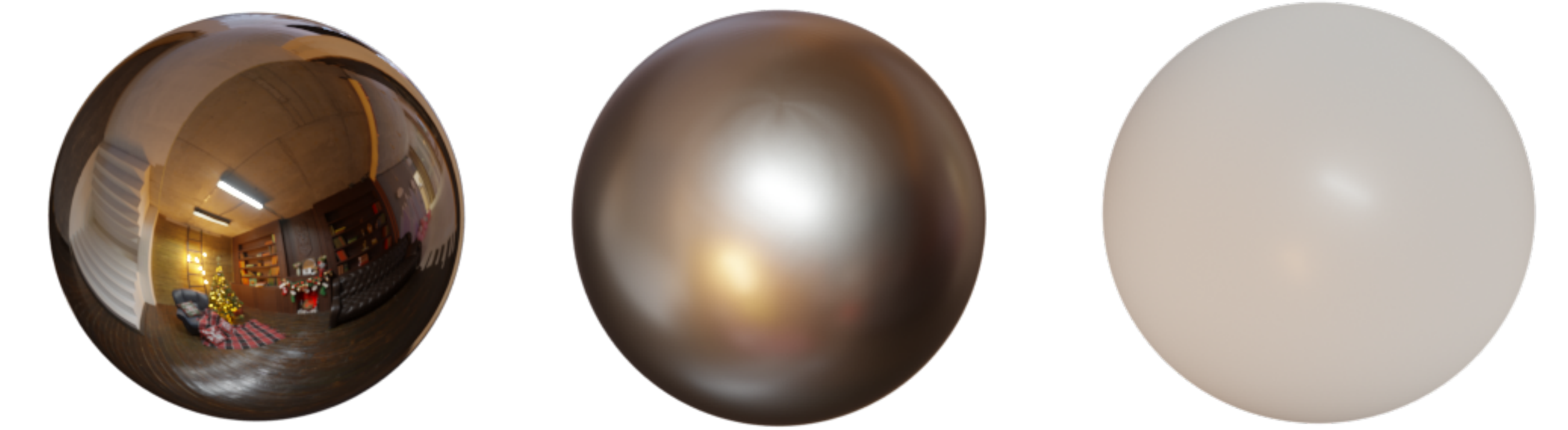}
  \caption{Three spheres with different materials: mirrored silver, matte silver, and diffuse grey.}\label{fig:sphere}
\end{wrapfigure}

\noindent
\textbf{Evaluation metrics.} Similar to Gardner2019 \cite{gardner2019deep} and EMlighting \cite{zhan2020emlight}, we use three spheres with different materials for evaluation, including mirror, matte silver, and gray diffuse, as illustrated in \textbf{Fig. \ref{fig:sphere}}. We render the three spheres by ground-truth HDR panoramas and the estimated illumination maps. Similar to \cite{zhan2020emlight}, we use Root Mean Square Error (RMSE), scale-invariant RMSE (si-RMSE), and Angular Error for evaluation. RMSE and si-RMSE measure pixel-level errors of lighting intensity. Angular Error measures the ratio of R, G, and B components. Different from \cite{zhan2020emlight}, we compute the metrics on the three rendered spheres and ignore the area of the rendered black background, which is invariant to the area of the background. Because there are no public codes for evaluation, we design our own hyper-parameters for the three spheres. We also use Fr\'{e}chet Inception Distance (FID)~\cite{heusel2017gans} to measure the distribution similarity between the ground-truth and the estimated illumination maps.

\subsection{Quantitative Evaluation}
\label{subsec:quantitative}
We compare StyleLight with three representative state-of-the-art illumination estimation methods, including a generation-based method \cite{zhan2020emlight} and two regression-based methods \cite{gardner2017learning,gardner2019deep}. Note that Gardner2017 \cite{gardner2017learning} used extra lighting annotations and Gardner2019 \cite{gardner2019deep} used extra depth annotations for training. StyleLight and EMlight \cite{zhan2020emlight} models do not require any extra annotation. For each method, we render three typical spheres (\textbf{Fig. \ref{fig:sphere}}) by the predicted illumination maps and ground-truth HDR panoramas on the test set and evaluate RMSE, si-RMSE, and Angular Error. We also evaluate FID to measure the distribution distance of ground-truth and predicted illumination maps. 

As shown in \textbf{Table \ref{tab:Comparisons}}, our StyleLight outperforms the compared methods on all metrics and materials. Among them, Gardner2017 and Gardner2019 focus on strong lighting sources and ignore low-intensity lighting, so they fail to achieve a desired RMSE and si-RMSE on the mirror sphere and do not preserve components of colors. EMlight first regresses a Gaussian map and fuses it with the LFOV image for illumination generation. The drawback is that the multi-step optimization could lead to sub-optimal performance. Since EMlight contains regression learning, it is difficult to train. EMlight\footnote{https://github.com/fnzhan/Illumination-Estimation} claimed that the model has to be trained on subsets of 100, 1000, 2500,... and the full set gradually and could collapse occasionally during training. We also see that our StyleLight largely outperforms other methods on FID, indicating that our predicted lighting distribution is closest to ground truth on a wide range of lighting. The results also suggest that StyleLight can synthesize plausible scenes and can achieve a promising result when rendering mirror materials. Our StyleLight also provides controllable lighting editing.

\begin{table*}[!htp]
\caption{Comparison with previous work. Gardner2017 and Gardner2019 are regression-based methods. EMlight is a generation-based method. Annotation denotes required manual annotations. The evaluation metrics include Angular Error, RMSE, si-RMSE and FID. M, S, and D denote mirror, matte silver, and diffuse material spheres, respectively. $\downarrow$ denotes that smaller is better.}\label{tab:Comparisons}
\begin{center}
\scalebox{0.9}{
\begin{tabular}{c|ccc|ccc|ccc|ccc}
\hline
\multirow{2}{2.5cm}{\centering\textbf{Methods}} &\multicolumn{3}{c|}{\textbf{Gardner2017 \cite{gardner2017learning}}} &\multicolumn{3}{c|}{\textbf{Gardner2019 \cite{gardner2019deep}}} &\multicolumn{3}{c|}{\textbf{EMlight \cite{zhan2020emlight}}}&\multicolumn{3}{c}{\textbf{StyleLight}}\\
\cline{2-13}
                            ~&M &S &D &M &S &D &M &S &D &M &S &D\\
\hline
Annotation & \multicolumn{3}{c|}{lighting sources} &\multicolumn{3}{c|}{depth}&\multicolumn{3}{c|}{No}&\multicolumn{3}{c}{No}\\
\hline
Angular Error$\downarrow$ & 6.68   &6.03 &5.47  &4.63 	&3.91	&3.24 &4.67 &3.47	&2.59 &\textbf{4.30}  	&\textbf{2.96}	&\textbf{2.41}\\
RMSE$\downarrow$ & 0.70    &0.38 &0.19  &0.85	&0.49	&- &0.59  	&0.33	&0.17 &\textbf{0.56} 	&\textbf{0.30}	&\textbf{0.15} \\
si-RMSE$\downarrow$& 0.65   &0.35  & 0.15  &0.70		&0.40	&0.14 &0.58	&0.31	& 0.12&\textbf{0.55}	&\textbf{0.29}	&\textbf{0.11}\\
\hline
\hline
FID$\downarrow$ & \multicolumn{3}{c|}{307.5} &\multicolumn{3}{c|}{344.1}&\multicolumn{3}{c|}{263.8}&\multicolumn{3}{c}{\textbf{137.7}}\\
\hline

\end{tabular}}
\end{center}
\end{table*}

\subsection{Qualitative Evaluation}
\label{subsec:qualitative}
In this section, we provide visual results to show the effectiveness of the proposed StyleLight model, including visual comparison with state-of-the-art lighting estimation methods and lighting editing.

\begin{figure*}[t]
  \centering
  \includegraphics[width=\linewidth]{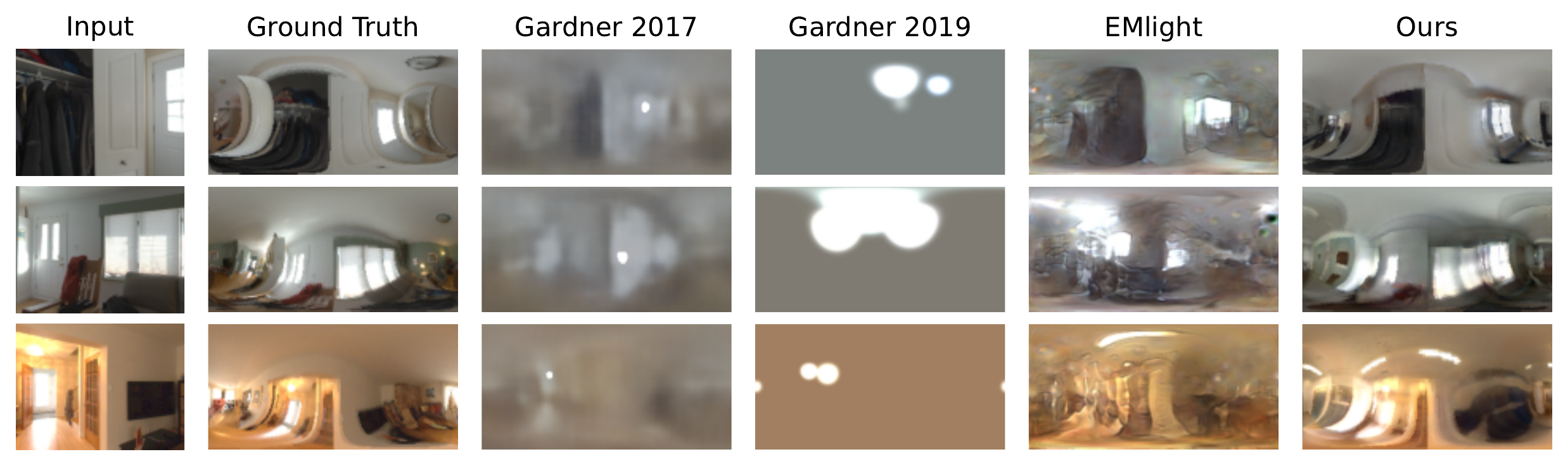}
  \caption{Visual comparisons of our method with several SOTA methods.}\label{fig:vis_comparision}
\end{figure*}

\begin{figure*}[ht]
  \centering
  \includegraphics[width=\linewidth]{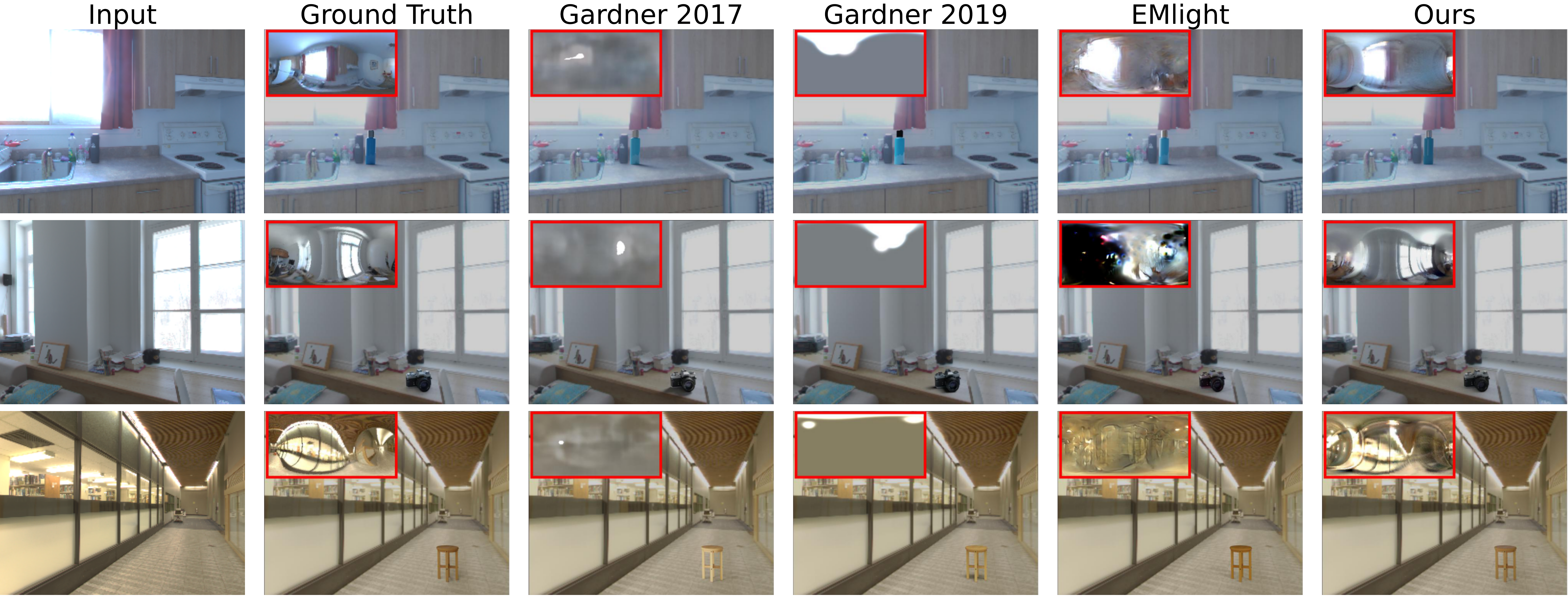}
  \caption{Visual comparisons on object insertion and lighting. For each input LDR LFOV image, the four different methods estimate illumination maps (at the top-left of LFOV images) for rendering newly inserted objects. }\label{fig:object_insertion}
\end{figure*}

\noindent
\textbf{Visual comparison with the state-of-the-art lighting estimation methods.} 
As depicted in \textbf{Fig. \ref{fig:vis_comparision}}, the predicted illumination maps by StyleLight are similar to the ground truth, and the proposed method obtains high-quality panoramas. Compared with the ground truth, Gardner2017 and Gardner2019 highlight strong lighting sources, which are not suitable to render mirror materials. EMlight considers both low-intensity and high-intensity lighting, but it obtains blurred images and it is difficult to train because of the unstable regression. Our StyleLight can synthesize panoramas of better quality, which can be further used for mirror rendering and lighting editing.

\noindent
\textbf{Visual comparison with the state of the art on object insertion and lighting.} We further compare StyleLight with state-of-the-art methods on object insertion and lighting. \textcolor{black}{The predicted HDR panorama is a sphere. The object is placed at the center of the panorama and a warping operator \cite{gardner2017learning} is used when inserted objects are not at the center of a panorama. Finally, image-based rendering is applied. The object insertion technique is similar to EMlight.} The four different methods estimate illumination maps for rendering newly inserted objects. As shown in \textbf{Fig. \ref{fig:object_insertion}}, compared with nearby objects in LFOV images, the inserted objects by Gardner2017 and Gardner2019 tend to lose low-intensity lighting information and generate inaccurate shadows. Thanks to the unique design of coupled dual-StyleGAN that integrates both HDR and LDR panoramas into a unified framework, StyleLight achieves a better visual outputs. 

\begin{figure*}[ht]
  \centering
  \includegraphics[width=0.9\linewidth]{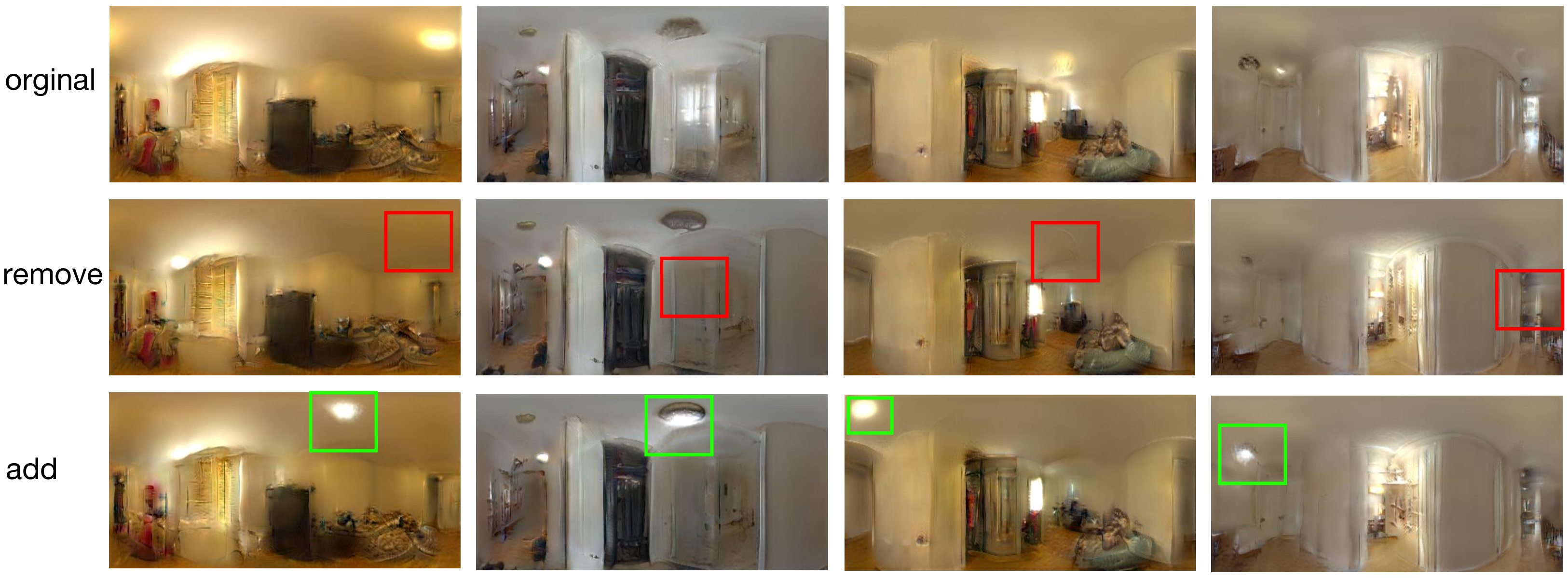}
  \caption{Lighting editing with StyleLight. For each original panorama (Row 1), StyleLight removes a light (red box, Row 2) and then adds a new light (green box, Row 3). }\label{fig:lighting_editing}
\end{figure*}

\noindent
\textbf{Lighting editing with StyleLight.} As shown in \textbf{Fig. \ref{fig:lighting_editing}}, given a panorama (the first column), we can remove lights in the second column and add new lights in the third column with the method presented in Sec.~\ref{subsec:light_edit}. 

\begin{figure*}[ht]
  \centering
  \includegraphics[width=0.85\linewidth]{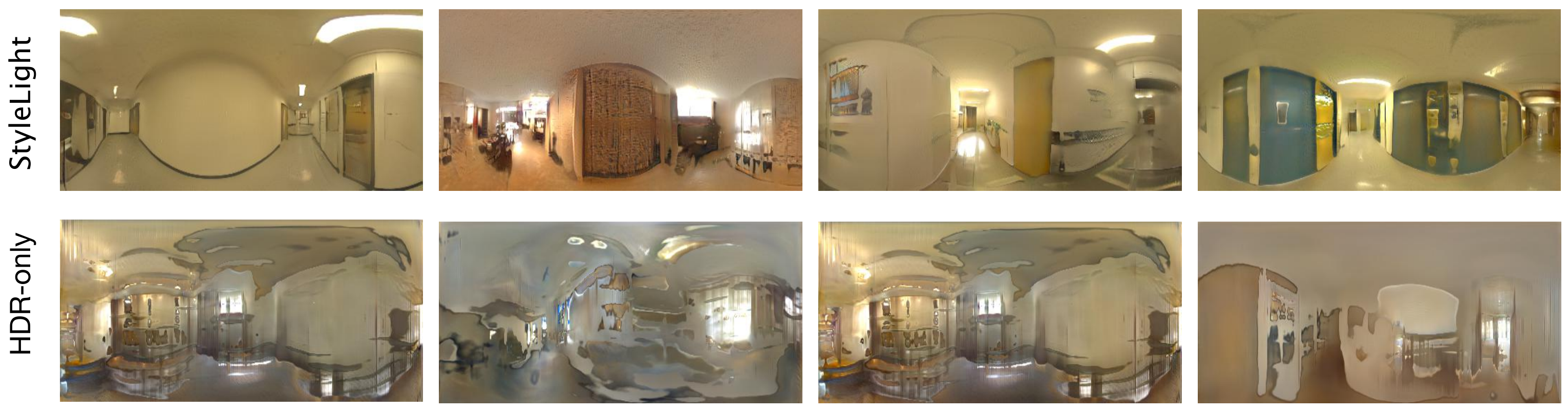}
  \caption{Effectiveness of the coupled dual-StyleGAN structure. The dual-StyleGAN generates both HDR and LDR panoramas. HDR-only directly generates HDR panoramas. }\label{fig:hdrldr_vs_hdr_only}
\end{figure*}

\noindent
\textbf{Effectiveness of the Coupled Dual-StyleGAN structure.}
We conduct an ablation study to analyze the effectiveness of the coupled dual-StyleGAN framework. As shown in \textbf{Fig. \ref{fig:hdrldr_vs_hdr_only}}, the first row shows the quality of StyleLight that uses two discriminators to learn LDR and HDR distributions. The second row shows the LDR panorama quality by using one discriminator to train HDR synthesis (denoted as HDR-only) and the LDR panoramas are computed by tone mapping. It is observed that LDR panoramas from the `HDR-only' suffer from severe distortion. This ablation demonstrates the effectiveness of the coupled dual-StyleGAN structure.

\begin{table}[ht]
\caption{Effectiveness of the focal-masked loss function. M, S, and D denote mirror, matte silver, and diffuse material spheres, respectively. $\downarrow$ denotes that smaller is better.}\label{tab:focal_masked}
 \centering
 \begin{tabular}{c|c|ccc|ccc}
 \hline
\multirow{2}{2.5cm}{\centering \textbf{ {metrics}}}&\textbf{baseline} [\textcolor{green}{20}]&\multicolumn{3}{c|}{\textbf{masked}}&\multicolumn{3}{c}{\textbf{focal-masked}}\\
\cline{2-8}
 ~& N/A & M & S & D & M & S & D \\
 \hline
   RMSE$\downarrow$& N/A & 0.57 & 0.31 & 0.16 & \textbf{0.56} & \textbf{0.30} & \textbf{0.15}\\
 si-RMSE$\downarrow$&  N/A & 0.56 & 0.30 & 0.12 & \textbf{0.55} & \textbf{0.29} & \textbf{0.11}\\
 \hline
 \end{tabular}
 \end{table}

\begin{wrapfigure}{r}{0.5\textwidth}
  \begin{center}
    \includegraphics[width=0.48\textwidth]{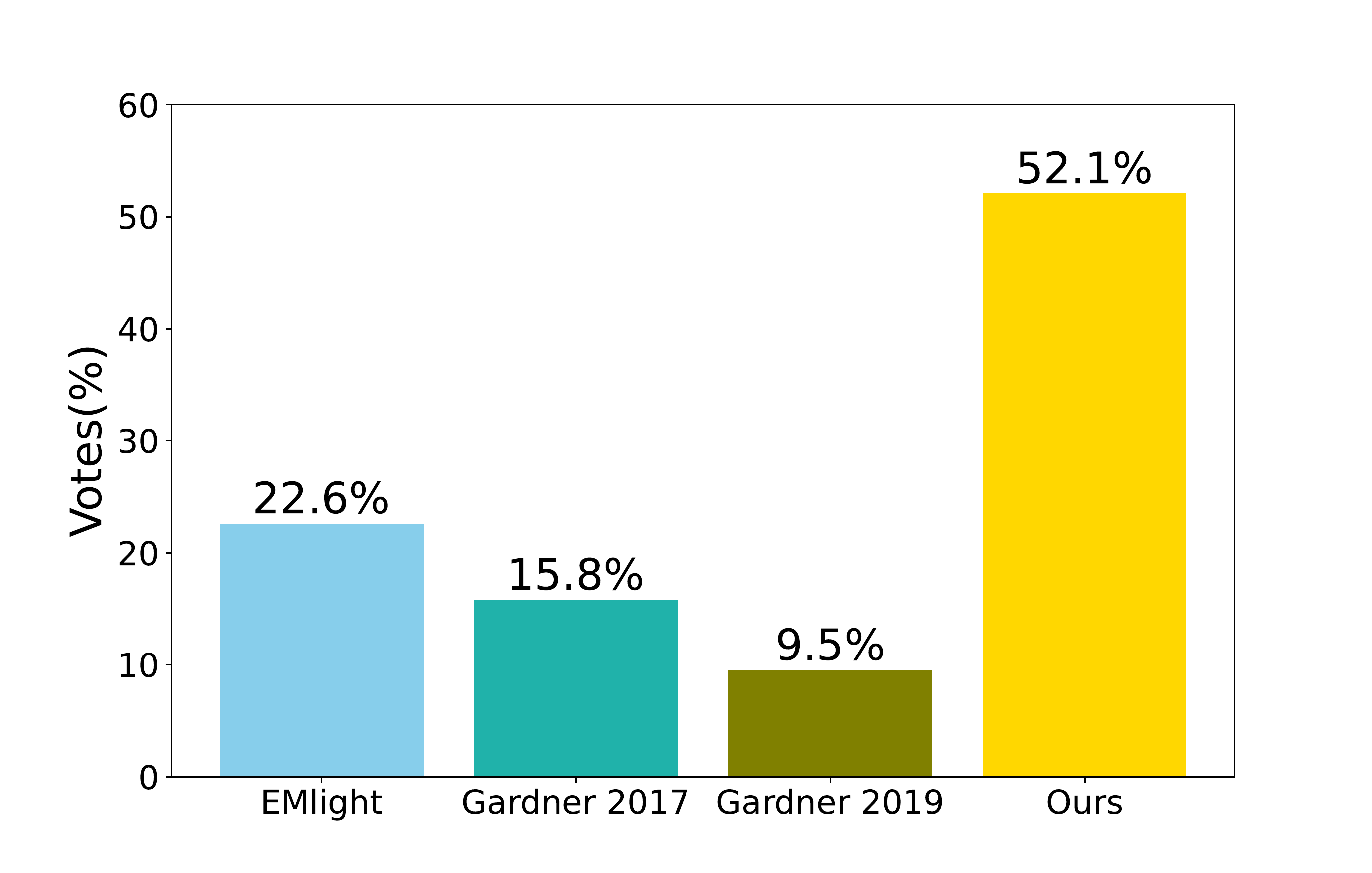}
  \end{center}
  \caption{User preference.}\label{fig:user_study}
\end{wrapfigure}

\noindent
\textcolor{black}{\textbf{Effectiveness of the focal-masked loss function.} We conduct an ablation study for the new loss in Eq. \ref{eq:gan_finetune_new}. We analyze RMSE and si-RMSE as shown in Table \ref{tab:focal_masked}. The existing GAN inversion is unavailable to LFOV images. We first develop a masked version and then extend it to a focal-masked version. The table shows the improvement of the proposed focal-masked loss function.}

\begin{figure*}[ht]
  \centering
  \includegraphics[width=\linewidth]{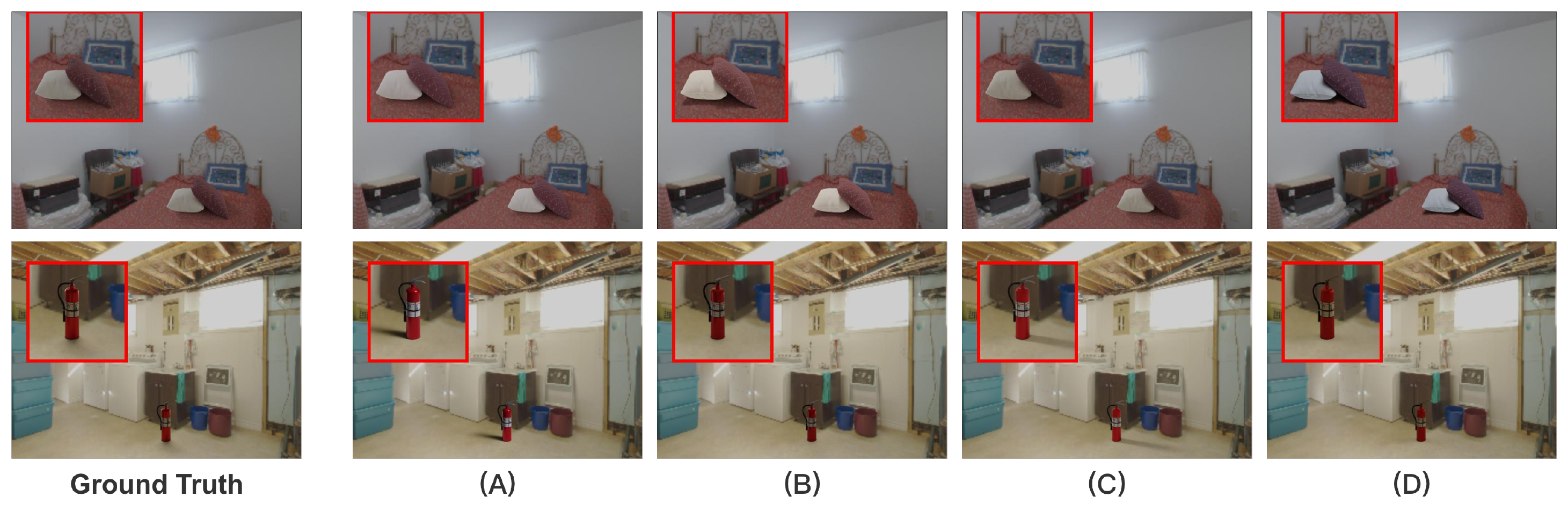}
  \caption{For each row, the inserted objects are rendered by HDR illumination maps estimated by four methods. Which one do you think is more closer to the object rendered by the ground-truth HDR illumination map? \\\begin{turn}{180}Second row: (A)Gardner 2019, (B)Ours, (C)Gardner 2017, (D)EMlight\end{turn}
  \\\begin{turn}{180} 
First row: (A)EMlight, (B)Gardner 2017, (C)Ours, (D)Gardner 2019;\end{turn}}\label{fig:user_study_example}
\end{figure*}

\noindent
\textbf{User Study.} We conduct a user study to analyze predicted illumination maps.  \textbf{Fig. \ref{fig:user_study_example}} shows two examples of the user study. \textcolor{black}{For each example, we insert an object into an image. We render the inserted object with illumination maps predicted by four methods}. Users were asked to pick the images that were more closer to the images rendered by ground-truth HDR panorama. We rendered different objects under 10 scenes, and 100 users participated in the study. Results are shown in \textbf{Fig. \ref{fig:user_study}}. We can observe that 52.1\% prefer our rendered results, which significantly outperform other methods.

\section{Conclusion}
\label{sec:conclusion}
We have presented a coupled dual-StyleGAN panorama synthesis network ({StyleLight}) for lighting estimation and editing. StyleLight offers high-quality lighting estimation that generates the full indoor panorama lighting from a single  LFOV image. With the trained StyleLight, we propose a structure-preserved lighting editing method that enables flexible lighting editing. Extensive experiments demonstrate that StyleLight achieves superior performance over state-of-the-art methods on indoor lighting estimation and also enables promising lighting editing on indoor HDR panoramas.

\noindent
\textcolor{black}{\textbf{Limitation.} StyleLight only considers indoor scenes in this paper. When applied to outdoor scenes, StyleLight needs a new set of hyper-parameters because lighting sources (e.g., area and intensity) are different between outdoor scenes and indoor scenes. An intuitive idea is to develop a generalizable method that adaptively tailors for various scenes, which is left to future research.}

\noindent
\textbf{Acknowledgement.} This work is supported by NTU NAP, MOE AcRF Tier 2 (T2EP20221-0033), and under the RIE2020 Industry Alignment Fund – Industry Collaboration Projects (IAF-ICP) Funding Initiative, as well as cash and in-kind contribution from the industry partner(s).

\clearpage
%
%
\bibliographystyle{splncs04}
\bibliography{egbib}

\clearpage

\section*{Supplementary}
\section*{A. More visual results}
\begin{figure}[h!]
  \centering
  \includegraphics[width=0.9\linewidth]{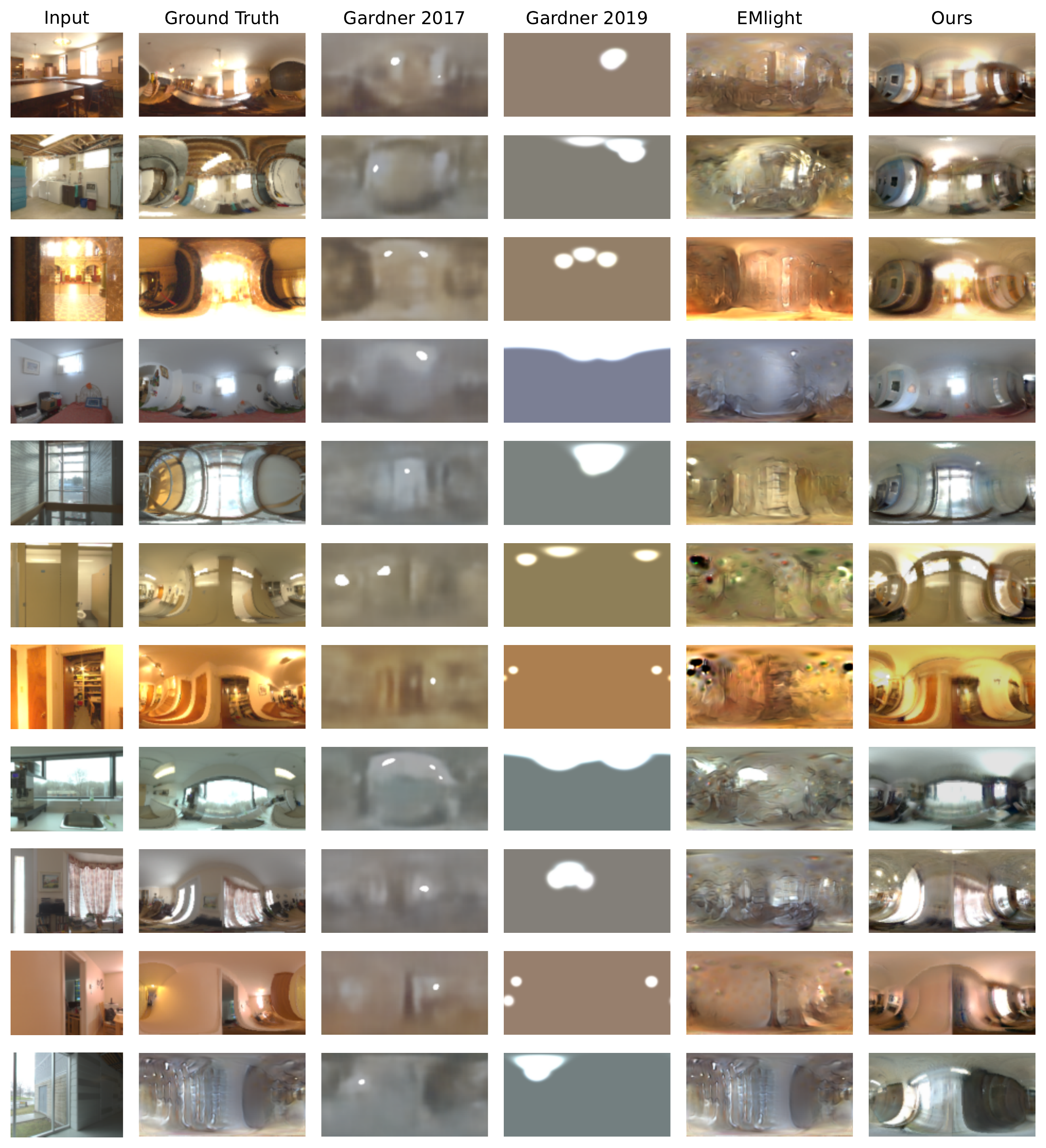}
  \caption{Visual comparisons of our method with several SOTA methods.}\label{fig:vis_comparision2}
\end{figure}

\begin{figure}[h!]
  \centering
  \includegraphics[width=0.95\linewidth]{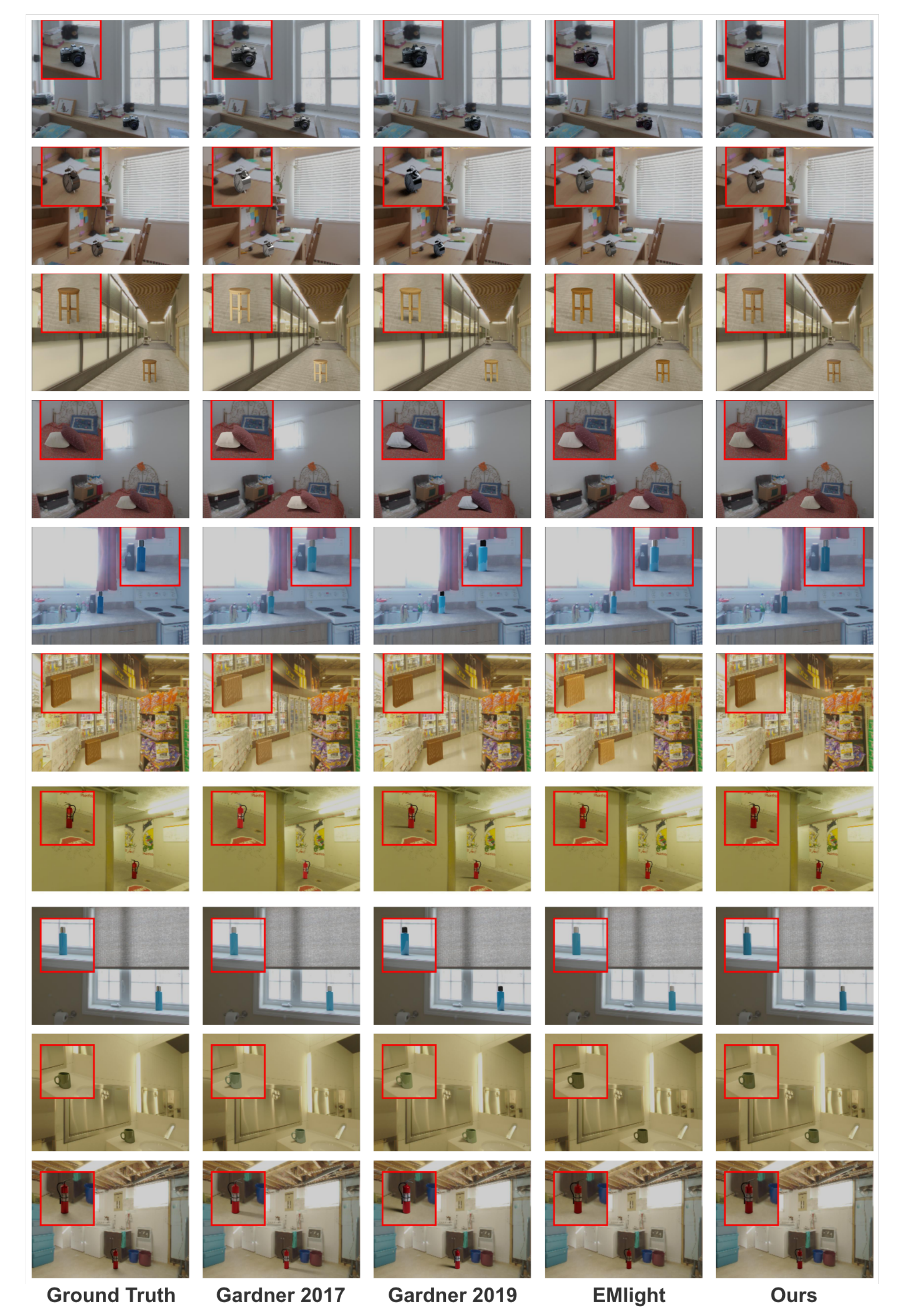}
  \caption{Visual comparisons on object insertion and lighting. For each input LDR LFOV image, the four different methods estimate illumination maps (at the top-left of LFOV images) for rendering newly inserted objects. }\label{fig:object_insertion2}
\end{figure}

In this section, we provide more visual results to show the effectiveness of the proposed StyleLight, as shown in Figs. \ref{fig:vis_comparision2} and \ref{fig:object_insertion2}.

\begin{figure}[h]
  \centering
  \includegraphics[width=\linewidth]{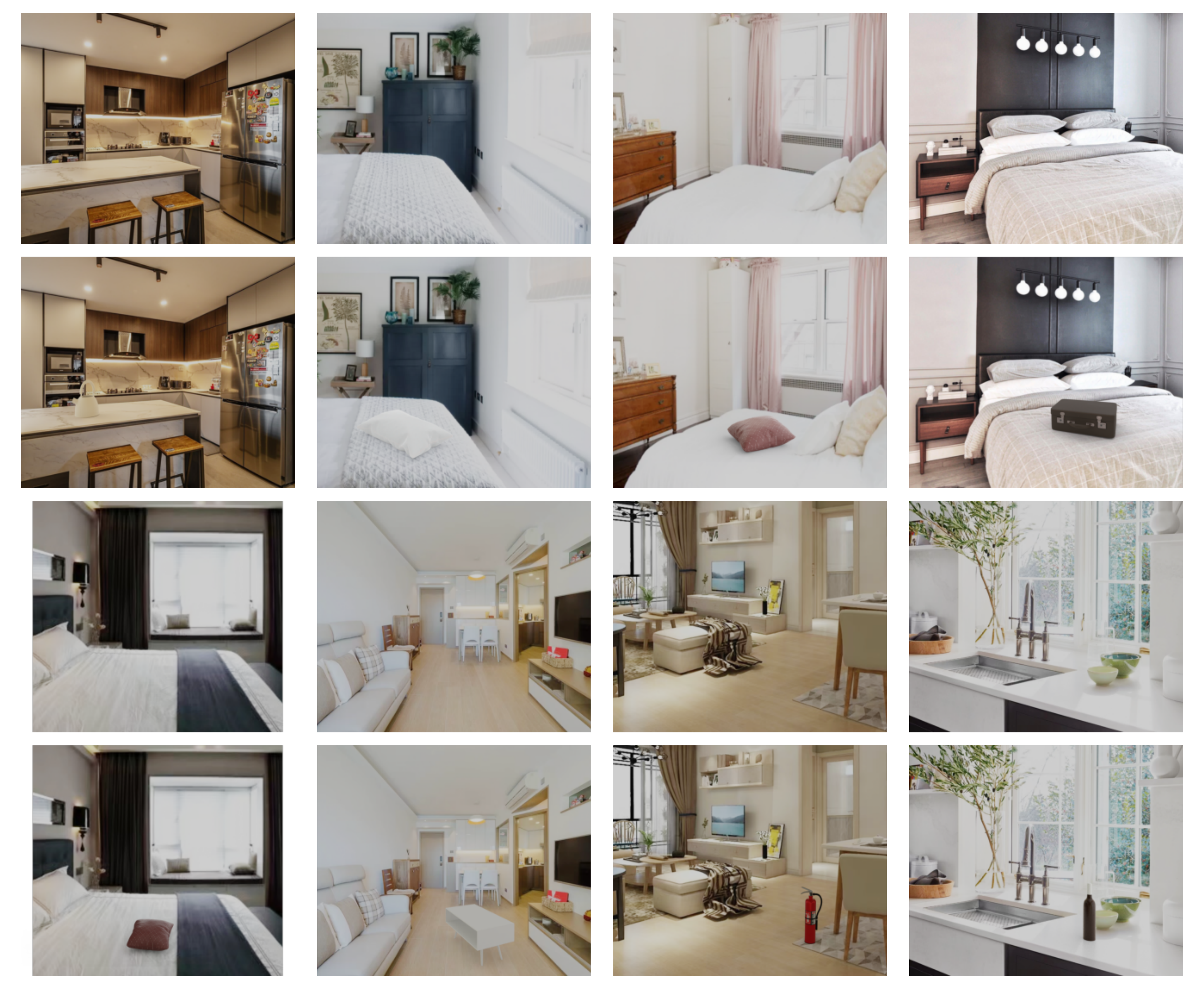}
  \vspace{-20pt}
  \caption{Object relighting on images from the internet. The four LDR LFOV images are from the internet. We apply StyleLight for lighting estimation on these images and render newly inserted objects. }\label{fig:figure_wild}
\end{figure}

\section*{B. Application on face relighting}
\begin{figure*}[t]
  \centering
  \includegraphics[width=\linewidth]{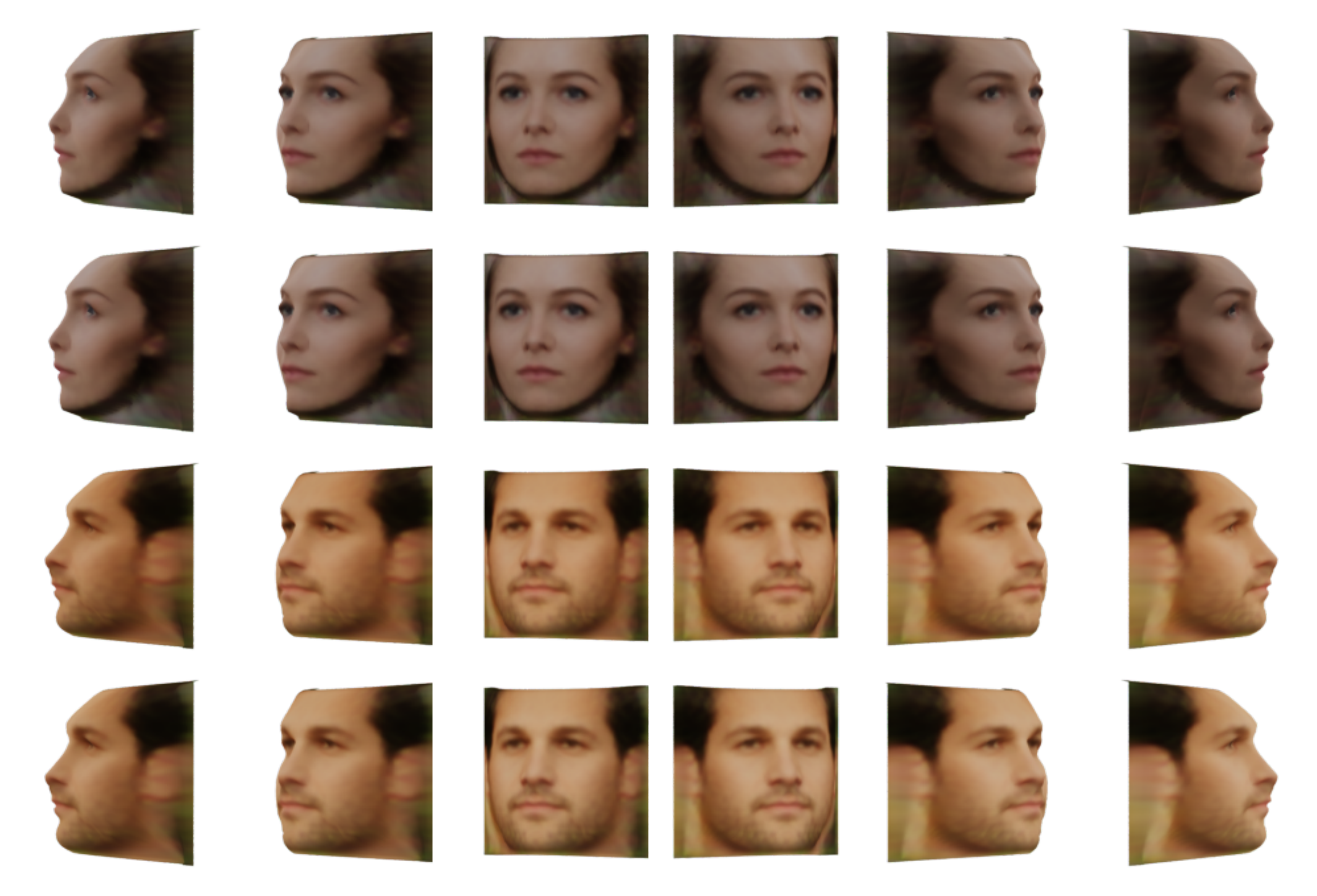}
  \vspace{-20pt}
  \caption{ \textbf{Comparisons on face rendering results.}  We show relighting results of human faces. The first and third rows are rendered by ground-truth HDR panorama; And the second and fourth rows are rendered by HDR panorama predicted by our StyleLight model.}\label{fig:face_lighting}
\end{figure*}

We apply StyleLight for face relighting. We use our predicted HDR panoramas and the corresponding ground truth to render human faces obtained by \cite{wu2020unsupervised}, as shown in \textbf{Fig. \ref{fig:face_lighting}}, we see that the rendered results of our predicted HDR panoramas are similar to the ground truth. 

\section*{C. Applications on the Internet's images.}
 We download some LFOV images from the Internet to evaluate the performance of object insertion in the wild. As shown in \textbf{Fig. \ref{fig:figure_wild}}, it is observed that the inserted objects seem realistic in wild scenes.

\section*{D. More details about implementation}
We provide more details about implementation in the experiments. In Equation \textcolor{red}{2}, we set $\lambda_{n}=1e5$. In Equation \textcolor{red}{3}, we set $\lambda_{L2}^{R}=10$. In Equation \textcolor{red}{4}, we set $\lambda_{L2}^{R'}=10$. $\eta=1$. In Equations \textcolor{red}{5} and \textcolor{red}{6}, we set $\beta_{L2}=10$. In Equation \textcolor{red}{8}, we set $\delta=1$ for decreasing lighting intensity and $\delta=-1$ for increasing lighting intensity.

\end{document}